\documentclass{article}

    \PassOptionsToPackage{numbers, compress}{natbib}

\usepackage[final]{neurips_2025}

\usepackage{graphicx}
\usepackage[utf8]{inputenc} 
\usepackage[T1]{fontenc}    
\usepackage{hyperref}       
\usepackage{url}            
\usepackage{booktabs}       
\usepackage{amsfonts}       
\usepackage{nicefrac}       
\usepackage{microtype}      
\usepackage{xcolor}         
\usepackage{wrapfig}
\usepackage{caption}
\usepackage{multirow}
\usepackage{booktabs}

\title{Self Distillation Fine-Tuning of Protein Language Models Improves Versatility in Protein Design}

\author{
 Amin Tavakoli \\
  Computing and Mathematical Sciences\\
  California Institute of Technology\\
  \texttt{amint@caltech.edu} \\
  \And
  Raswanth Murugan\\
  Department of Computer Science\\
  The City University of New York\\
  \texttt{raswanth99@gmail.com} \\
  \And
  Ozan Gokdemir\\
  Department of Computer Science\\
  University of Chicago\\
  \texttt{ogokdemir@uchicago.edu} \\
  \And
  Arvind Ramanathan\\
  Data Science and Learning Division\\
  Argonne National Laboratory\\
  \texttt{ramanathana@anl.gov} \\
  \And
  Frances Arnold\\
  Division of Chemistry and Chemical Engineering\\
  California Institute of Technology\\
  \texttt{frances@cheme.caltech.edu} \\
  \And
  Anima Anandkumar\\
  Computing and Mathematical Sciences\\
  California Institute of Technology\\
  \texttt{anima@caltech.edu} \\
}

\begin{document}

\maketitle

\begin{abstract}

Supervised fine-tuning (SFT) is a standard approach for adapting large language models to specialized domains, yet its application to protein sequence modeling and protein language models (PLMs) remains ad hoc. This is in part because high-quality annotated data are far more difficult to obtain for proteins than for natural language. We present a simple and general recipe for fast SFT of PLMs, designed to improve the fidelity, reliability, and novelty of generated protein sequences. 
Unlike existing approaches that require costly precompiled experimental datasets for SFT, our method leverages the PLM itself, integrating a lightweight curation pipeline with domain-specific filters to construct high-quality training data. These filters can independently refine a PLM’s output and identify candidates for \textit{in vitro} evaluation; when combined with SFT, they enable PLMs to generate more stable and functional enzymes, while expanding exploration into protein sequence space beyond natural variants.
Although our approach is agnostic to both the choice of protein language model (PLM) and the protein system, we demonstrate its effectiveness with a genome-scale PLM (GenSLM) applied to the tryptophan synthase enzyme family. The supervised fine-tuned model generates sequences that are not only more novel but also display improved characteristics across both targeted design constraints and emergent protein property measures.
\end{abstract}
\section{Introduction}\label{sec:intro}

Large protein language models (PLMs) have become a crucial tool in computational protein design by offering scalable generative frameworks for exploring protein sequence space. PLMs such as ESM \cite{lin2023evolutionary}, ProtGPT2 \cite{ferruz2022protgpt2}, ProGen \cite{madani2020progen}, and GenSLM \cite{zvyagin2023genslms} are pretrained on millions of natural protein sequences and have shown that the transformer architecture can capture evolutionary constraints and structural regularities that underpin protein functionality. These models can generate vast numbers of unseen sequences in seconds, which accelerates the design process far beyond what is feasible with traditional experimental or simulation-based methods.
Yet, important considerations need to be addressed regarding the practical utility of PLMs. First, the majority of generated sequences are not functionally viable, requiring additional mechanisms to prioritize promising candidates \cite{bhethanabotlacapturing, lambert2025sequence}. Second, pretrained PLMs are typically optimized for broad next-token prediction objectives, and do not directly align with downstream design goals such as stability, catalytic activity, or specificity. Addressing these challenges requires methods to adapt pretrained PLMs toward targeted objectives while preserving their generalization ability.

Supervised fine-tuning (SFT) offers a principled way to adapt pretrained protein language models (PLMs) to specific design objectives \cite{wang2025supervised, sledzieski2024democratizing}. By further training on curated datasets drawn from defined protein families and experimentally validated variants, SFT shifts model priors toward sequences consistent with structural and functional constraints.
In large language models (LLMs) SFT has become the standard recipe for domain adaptation and task alignment \cite{shao2024deepseekmath, rafailov2023direct}, often serving as a precursor to reinforcement learning approaches \cite{luo2024robustft}. In natural language, SFT is performed via self-distillation where synthetic data is generated by the model itself. 
In contrast, applications of SFT in protein modeling have remained inconsistent, with limited decisions in data selection, ad hoc filtering, and evaluation \cite{schmirler2024fine, sledzieski2024democratizing}, largely due to the difficulty of obtaining high-quality experimentally labeled data.
Such variability obscures reproducibility and limits the systematic integration of SFT and its benefits into protein design pipelines.

A central challenge underlying this inconsistency is the reliance of SFT on large-scale, high-quality protein sequence annotations. While natural language annotations can be generated rapidly by non-experts, evaluating protein functionality requires labor-intensive and costly experimental assays. This fundamental barrier constrains the scalability of SFT for PLMs, particularly when exploring novel protein families for which annotated data are sparse or unavailable. Computational proxies for annotation can potentially address this bottleneck, but their inherent inaccuracies further complicate training, as they cannot guarantee the identification of functionally viable sequences.

We propose a simple and general recipe for filtering functionally viable artificially generated protein sequences to enable self-distilling SFT. The framework emphasizes standardized computational data curation, the integration of lightweight sequence- and structure-based filters, and reproducible filtering procedures that align with the natural criteria of protein variants' viability. 
The filters are broadly applicable across PLMs and protein families for (1) purifying final candidates for \textit{in vitro} analysis; and (2) curating SFT data. The design of our filters allows flexible control over similarity measures by sampling from a wide range of proteins. This enables tuning the trade-off between sequence novelty and functionality—two essential considerations for evaluating PLM performance in protein design. To illustrate the approach, we focus on the $\beta$-chain of the tryptophan synthase complex (TrpB), a well-studied enzyme in amino acid biosynthesis. We fine-tune a 25M-parameter GenSLM model on TrpB sequences using filtered data and evaluate the resulting model on both explicitly enforced properties—including sequence length, active-site conservation, and predicted folding confidence (pLDDT)—and emergent properties not directly optimized during filtering, such as predicted stability and docking scores to natural substrates. This case study demonstrates that filtering-based SFT improves not only targeted properties but also generalizes to unseen objectives, underscoring the broader utility of our recipe for protein engineering.

\section{Materials and Methods}
Here, we describe the protein language model (PLM), and the filtering pipeline used to curate high-quality data for supervised fine-tuning (SFT). Details on the studied protein family are presented in the Appendix. As discussed above, SFT improves the fidelity of generated sequences while also enhancing their novelty. Both aspects are critical for generative protein design: higher fidelity increases the likelihood of functional proteins, while greater novelty allows exploration beyond natural sequence space, potentially uncovering proteins with unseen properties that can serve as valuable starting points for further analysis, such as directed evolution \cite{Arnold2017DirectedEB}. However, the trade-off between novelty and fidelity must be carefully managed. Excessive emphasis on novelty can compromise structural plausibility and yield non-functional sequences. To investigate this balance, we evaluate different SFT strategies that differ in how they sample from regions of varying sequence identity, thereby probing distinct regimes of the fidelity–novelty spectrum.

\begin{figure}[h]
\begin{center}
\centerline{\includegraphics[width=0.9\columnwidth]{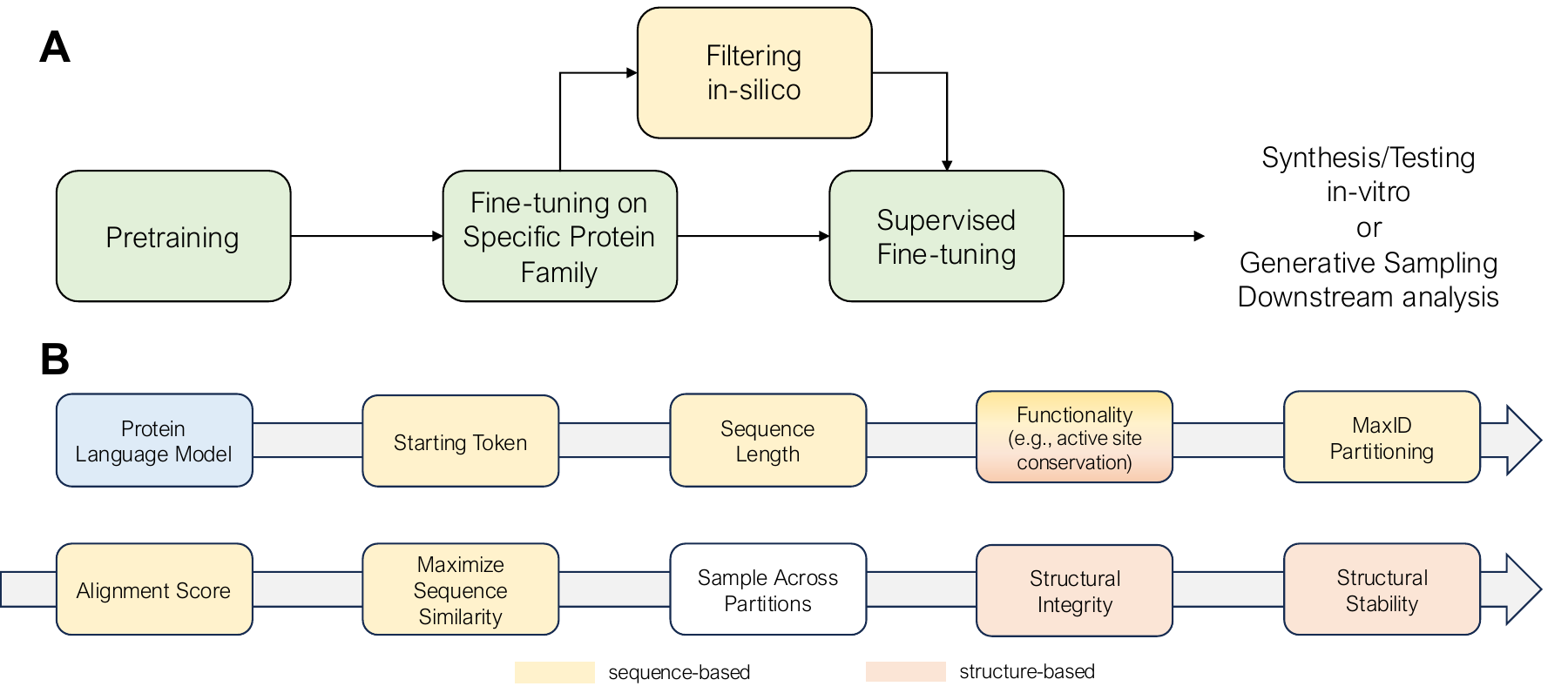}}
\caption{A: The general framework of the supervised fine-tuning pipeline using the artificially generated sequences. B: The sequence of the filters used to curate data for SFT.}
\label{fig:main}
\end{center}
\vspace{-3mm}
\end{figure}

\subsection{Protein Language Model}

We use Genome-Scale Language Model (GenSLM) \cite{zvyagin2023genslms}, a genome-scale transformer-based  \cite{vaswani2017attention} language model, first designed to capture long-range dependencies and predict the evolutionary dynamics of SARS-CoV-2 \cite{cosar2022sars}. GenSLMs include models ranging from 25M to 25B parameters. All variants were pretrained on 110 million prokaryotic genes from the Bacterial and Viral Bioinformatics Resource Center (BV-BRC), followed by fine-tuning on 1.5 million SARS-CoV-2 genomes to specialize in viral evolutionary patterns. A key distinction of GenSLM is its codon level tokenization, where DNA sequences are represented as triplets of nucleotides (64 codons) rather than individual bases. This representation mirrors the central dogma, directly reflecting the translation of DNA to protein, and allows the model to capture synonymous substitutions and protein-level functional effects within its learned representations.



\subsubsection{Fine-tuning on Tryptophan Synthase Family}
The protein system used in this study is the beta subunit of tryptophan synthase enzyme, TrpB (see Section 6.1 in the Appendix). We collect a fine-tuning dataset containing TrpB gene sequences from BV-BRC using the query parameters gene name: trpB and EC: 4.2.1.20, yielding 30,000 unique trpB nucleotide sequences. Fine-tuning was performed for 10 epochs with full parameter updates, using the standard autoregressive cross-entropy objective at the codon level. No diffusion-based hierarchical modeling or reward-guided generation of GenSLM is used in this stage. 
The fine-tuned model is used to generate a batch of 60,000 TrpB sequences used for the filtering stages.

\section{Filtering Pipeline}
Here, we outline the sequence of filters applied to 60,000 TrpB sequences generated by the fine-tuned GenSLM. The filters are discriminative, sequentially removing the least viable sequences at each stage. They are also designed to balance sequence novelty and functionality, ensuring that final candidates remain biochemically plausible while minimizing similarity to natural variants. This is achieved by continuously comparing generated sequences to a reference set of functional TrpB enzymes (see Section 6.2 of the Appendix for details).

A schematic overview of the filtering process is shown in Figure \ref{fig:main} B. The filters are applied in a hierarchical order to optimize computational efficiency. Lightweight sequence-level checks, such as the starting-token filter, are applied first to the full set of generated sequences. More computationally intensive steps, such as pLDDT-based structure confidence prediction, are deferred to the final stage and applied to a smaller set of sequences that have passed all prior filters. The final set of selected candidates will be used as novel and reliable sequences to perform SFT on GenSLM.

\textbf{Starting Token.} Because GenSLM is autoregressive, the starting token strongly influences the quality of the entire sequence. We retain only sequences beginning with the ATG codon or methionine (\textit{M} in amino acid representation), consistent with natural TrpB translation initiation.

\textbf{Sequence Length.} Sequence length is tightly coupled to protein structure and function. For TrpB, natural variants fall within a narrow range, with an average length of $363.6 \pm 57.9$ AA residues in the reference set. We restrict candidates to lengths within two standard deviation of this mean.

\textbf{Active Site Conservation.} To preserve functional plausibility, active-site conservation can be enforced at either the structural or sequence level. At the structural level, one could align ESMFold-predicted models to reference structures and retain only candidates that preserve catalytic-site geometry. In this work, we applied the sequence-based approach: generated sequences were aligned to a reference TrpB (pfTrpB, PDB: 5DW0), where residue Lys82 defines the catalytic lysine, and only sequences conserving this residue were retained. This important step efficiently eliminates catalytically implausible sequences, even if they appear structurally consistent.

\textbf{Maximum Sequence Identity (Max ID).} For the assessment and control of the sequence novelty of the generated sequences, we compute the maximum sequence identity (max ID) of each candidate against all reference sequences, following \cite{madani2020progen}. The distribution of max IDs reveals distinct modes of generation quality, ranging from highly novel to near-identical variants.

\textbf{Partitioning.} To ensure balanced exploration of sequence space, sequences are partitioned into max ID bins (40--50\%, 50--60\%, \dots, 90--100\%) to ensure diverse sampling. This prevents over-representation of highly similar sequences in the SFT data and maintain the coverage of more novel candidates (Figure \ref{fig:partitions}).

\textbf{Alignment Score.} To promote functionality in the final filtered set, sequences are ranked within each partition by an alignment score, calculated as a weighted combination of global and local pairwise alignments against the reference set. To further ensure diversity, sequences with high pairwise similarity (identified using BLAST) are removed, preventing near-duplicates from being retained.

\textbf{Sampling Strategy.} From each partition, the top-$N$ sequences ranked by alignment score are retained. The choice of $N$ controls the trade-off between the sequence novelty and functionality in the final set, and thus influences the properties of generated enzymes using the fine-tuned model. To illustrate this effect, we employ two sampling strategies: (1) \textbf{\textit{Functionality}:} samples are drawn primarily from the most similar partition (90--100\% maxID) to encourage functionality, and (2) \textbf{\textit{Novelty}:} samples are drawn more uniformly across all partitions of maximum identity. In both cases, the combined filters ensure that only high-quality sequences are retained.

\textbf{Structural Integrity.} We next assess the structural integrity of the sequences using ESMFold predictions. Sequences with predicted structures that have the average pLDDT scores below 80\% are discarded, ensuring only structurally confident models are retained. This step substantially enriches sequences closer to natural TrpB foldability.

\textbf{Stability.} Finally, structural stability can be assessed using Rosetta Relax energies \cite{rohl2004protein}. Although correlated with folding confidence, this metric provides an additional check on thermodynamic plausibility. Although this is an important part of the general filtering recipe, we chose not to include it as a filter in order to keep the pipeline lightweight and to evaluate the fine-tuned model for stability as an unseen property.

\begin{table}[h]
\centering
\small
\caption{Statistics for each filtering step and the number of final candidates selected for the SFT.}
\begin{tabular}{lcc}
\toprule
Filtering Step & Input Sequences (\#) & Filtered Out (\%) \\
\midrule
Starting Token (M/ATG) & 60000 & 0.016 \\
Sequence Length & 59990 & 6.66 \\
Active Site Conservation & 55990 & 1.76 \\
Max ID + Alignment Score  & 55000 & 45.03 \\
Sampling Strategy (Novelty - Functionality)& 20320 - 7110 & 63.06 - 87.08\\
Structural Integrity & 20320 - 7110 & 1.63 \\
\midrule
Final Retained & 20000 - 7000 \\
\bottomrule
\label{tab:stats}
\end{tabular}
\end{table}

\section{Results}
Starting from 60,000 generated TrpB sequences, we applied the filtering pipeline followed by two sampling strategies: (1) \textbf{Functionality}, yielding $\sim$7,000 sequences with high similarity to natural variants; and (2) \textbf{Novelty}, yielding $\sim$20,000 sequences with greater diversity across partitions of maximum identity. The number of filtered sequences at each stage of the pipeline is shown in Table \ref{tab:stats}. Supervised fine-tuning of the 25M-parameter GenSLM was then performed for three epochs with a batch size of eight, a learning rate of $2 \times 10^{-5}$, and the Adam optimizer, using next-token prediction with cross-entropy loss over the curated TrpB sequences.  

To evaluate the effect of the self-distillation fine-tuning, both fine-tuned models and the reference (base) model are used to generate 1,000 sequences. Each sequence was translated into its amino acid representation and evaluated on the targeted properties: sequence length, pLDDT score, and active-site conservation; and the unseen properties: stability and functionality. For the case of TrpB, we quantify one measure of functionality via the binding affinity of PLP (the enzyme’s cofactor) in the active site defined in the vicinity of the catalytic lysine. The improvement results of the targeted properties are presented in Table \ref{tab:evaluation}, whereas the results of the unseen properties are shown in Figure \ref{fig:main}. Although both sets of properties—unseen and targeted—improved, as expected, the enhancement in targeted properties was more pronounced.

\begin{figure*}[h]  
\centering
\includegraphics[width=0.48\textwidth]{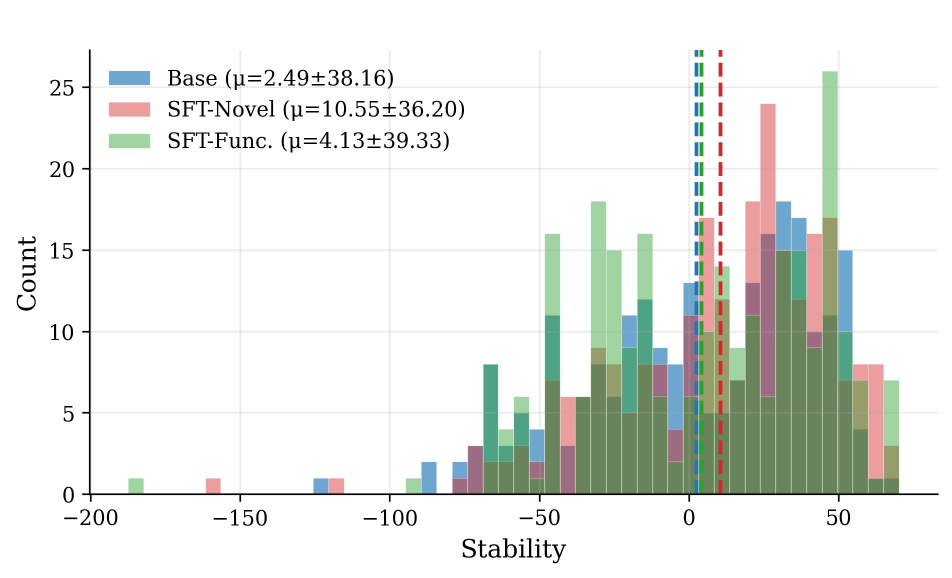}
\hfill
\includegraphics[width=0.48\textwidth]{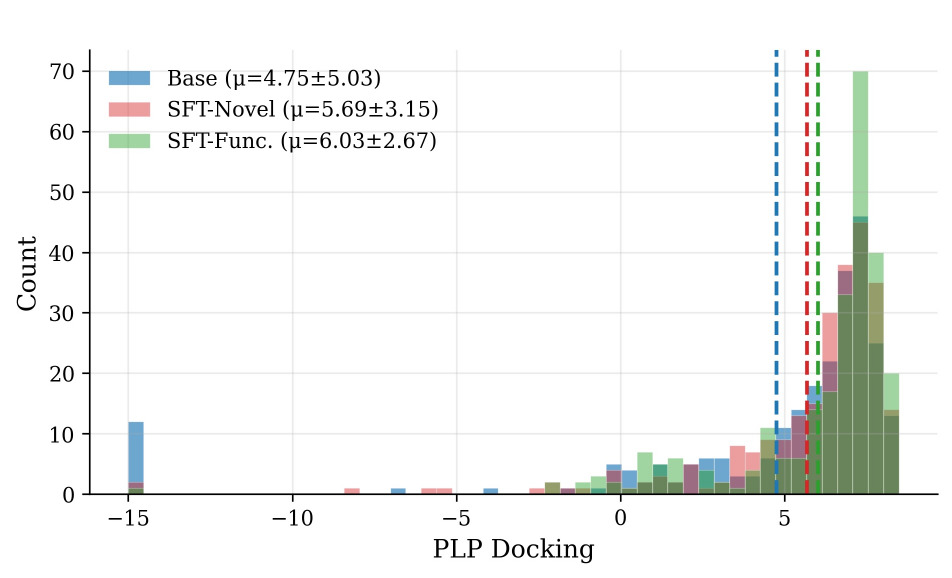}
\caption{(a) Stability scores. (b) PLP docking scores. In both cases, the plots show reversed magnitude, so higher values are better. The distribution shift is observed in both structural stability and the PLP Docking scores of generated samples. post SFT training}
\label{fig:main}
\end{figure*}

\begin{table}[h]
\centering
\small
\caption{Comparison of targeted property metrics between the reference and SFT models. After SFT, the model shows near-universal compliance with sequence length constraints, active-site preservation, and high-confidence ESM-based structure predictions.}
\begin{tabular}{lcc}
\toprule
 & Reference Model & SFT Model \\
\midrule
Sequence Length ($L \in [386, 430]$) & 93\% & 99.9\% \\
Active Site Conservation Rate & 95\% & 99.2\%-99.6\% \\
pLDDT (threshold = 0.8) & 97\% & 99.9\% \\
\bottomrule
\label{tab:evaluation}
\end{tabular}
\end{table}

\begin{figure}[h]
\begin{center}
    \includegraphics[width=0.6\columnwidth]{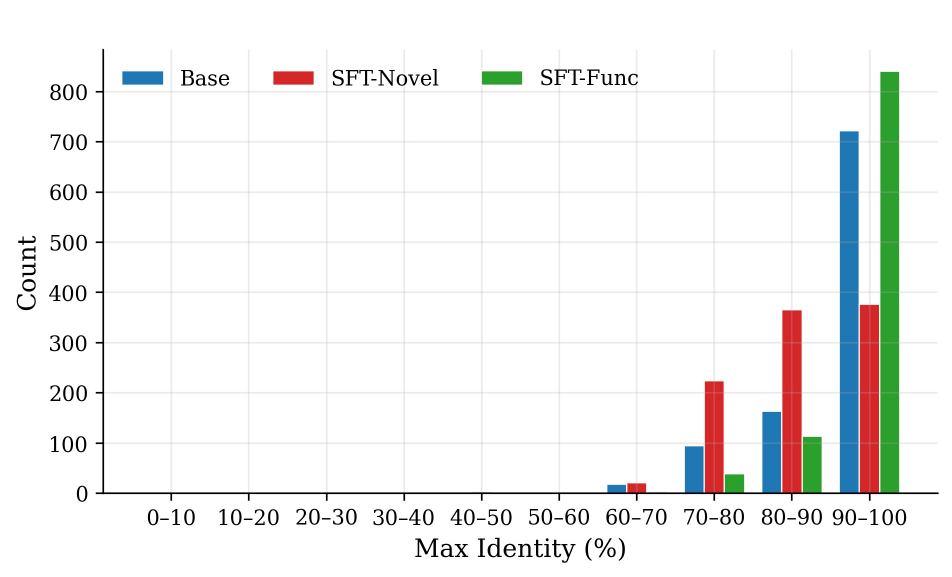}
    \caption{SFT-Novel generates more sequences in lower maxID regions compared to the vanilla GenSLM model, resulting in greater sequence diversity.}
    \label{fig:partitions}
\end{center}
\end{figure}

As shown in Figure \ref{fig:main}, sequences generated by the fine-tuned models exhibit enhanced stability and PLP binding affinity. We also observe that sequences fine-tuned with the functionality-focused strategy show larger improvements in functional performance. This trend is consistent with the maximum sequence identity of samples generated by both fine-tuned models, as illustrated in Figure \ref{fig:partitions}. Specifically, the model fine-tuned for novelty produces sequences with greater divergence from natural TrpBs, whereas the functionality-focused SFT generates sequences that are more similar to natural variants but exhibit higher functional performance.

\section{Conclusion}

We developed a set of computational filters to refine artificially generated protein sequences for self-distillation supervised fine-tuning (SFT) of protein language models (PLMs). SFT on PLMs is typically challenging and slow due to the scarcity of high-quality training data. Our results demonstrate that SFT using sequences curated by these filters improves the quality of model-generated sequences across multiple aspects, including stability, functionality, and sequence novelty, without relying on external datasets or wet-lab experiments. Furthermore, by applying more stringent thresholds, the filters can prioritize protein candidates for \textit{in vitro} testing, streamlining the identification of functional proteins for experimental validation.

\bibliographystyle{unsrt}
\bibliography{main}

@article{zvyagin2023genslms,
  title={GenSLMs: Genome-scale language models reveal SARS-CoV-2 evolutionary dynamics},
  author={Zvyagin, Maxim and Brace, Alexander and Hippe, Kyle and Deng, Yuntian and Zhang, Bin and Bohorquez, Cindy Orozco and Clyde, Austin and Kale, Bharat and Perez-Rivera, Danilo and Ma, Heng and others},
  journal={The International Journal of High Performance Computing Applications},
  volume={37},
  number={6},
  pages={683--705},
  year={2023},
  publisher={SAGE Publications Sage UK: London, England}
}

@article{cosar2022sars,
  title={SARS-CoV-2 mutations and their viral variants},
  author={Cosar, Begum and Karagulleoglu, Zeynep Yagmur and Unal, Sinan and Ince, Ahmet Turan and Uncuoglu, Dilruba Beyza and Tuncer, Gizem and Kilinc, Bugrahan Regaip and Ozkan, Yunus Emre and Ozkoc, Hikmet Ceyda and Demir, Ibrahim Naki and others},
  journal={Cytokine \& growth factor reviews},
  volume={63},
  pages={10--22},
  year={2022},
  publisher={Elsevier}
}

@article{madani2020progen,
  title={Progen: Language modeling for protein generation},
  author={Madani, Ali and McCann, Bryan and Naik, Nikhil and Keskar, Nitish Shirish and Anand, Namrata and Eguchi, Raphael R and Huang, Po-Ssu and Socher, Richard},
  journal={arXiv preprint arXiv:2004.03497},
  year={2020}
}

@article{vaswani2017attention,
  title={Attention is all you need},
  author={Vaswani, Ashish and Shazeer, Noam and Parmar, Niki and Uszkoreit, Jakob and Jones, Llion and Gomez, Aidan N and Kaiser, {\L}ukasz and Polosukhin, Illia},
  journal={Advances in neural information processing systems},
  volume={30},
  year={2017}
}

@article{pickett2012vipr,
  title={ViPR: an open bioinformatics database and analysis resource for virology research},
  author={Pickett, Brett E and Sadat, Eva L and Zhang, Yun and Noronha, Jyothi M and Squires, R Burke and Hunt, Victoria and Liu, Mengya and Kumar, Sanjeev and Zaremba, Sam and Gu, Zhiping and others},
  journal={Nucleic acids research},
  volume={40},
  number={D1},
  pages={D593--D598},
  year={2012},
  publisher={Oxford University Press}
}

@book{gore2000spectrophotometry,
  title={Spectrophotometry and spectrofluorimetry: a practical approach},
  author={Gore, Mike},
  volume={225},
  year={2000},
  publisher={OUP Oxford}
}

@article{tsao2003polyphenolic,
  title={Polyphenolic profiles in eight apple cultivars using high-performance liquid chromatography (HPLC)},
  author={Tsao, Rong and Yang, Raymond and Young, J Christopher and Zhu, Honghui},
  journal={Journal of agricultural and food chemistry},
  volume={51},
  number={21},
  pages={6347--6353},
  year={2003},
  publisher={ACS Publications}
}

@incollection{rohl2004protein,
  title={Protein structure prediction using Rosetta},
  author={Rohl, Carol A and Strauss, Charlie EM and Misura, Kira MS and Baker, David},
  booktitle={Methods in enzymology},
  volume={383},
  pages={66--93},
  year={2004},
  publisher={Elsevier}
}

@article{lin2023evolutionary,
  title={Evolutionary-scale prediction of atomic-level protein structure with a language model},
  author={Lin, Zeming and Akin, Halil and Rao, Roshan and Hie, Brian and Zhu, Zhongkai and Lu, Wenting and Smetanin, Nikita and Verkuil, Robert and Kabeli, Ori and Shmueli, Yaniv and others},
  journal={Science},
  volume={379},
  number={6637},
  pages={1123--1130},
  year={2023},
  publisher={American Association for the Advancement of Science}
}

@article{uniprot2023,
  author = {UniProt Consortium},
  title = {UniProt: the Universal Protein Knowledgebase in 2023},
  journal = {Nucleic Acids Research},
  year = {2023},
  volume = {51},
  pages = {D523--D530},
  doi = {10.1093/nar/gkac1052},
  url = {https://doi.org/10.1093/nar/gkac1052}
}

@article{ferruz2022protgpt2,
  title={ProtGPT2 is a deep unsupervised language model for protein design},
  author={Ferruz, Noelia and Schmidt, Steffen and H{\"o}cker, Birte},
  journal={Nature communications},
  volume={13},
  number={1},
  pages={4348},
  year={2022},
  publisher={Nature Publishing Group UK London}
}

@article{shao2024deepseekmath,
  title={Deepseekmath: Pushing the limits of mathematical reasoning in open language models},
  author={Shao, Zhihong and Wang, Peiyi and Zhu, Qihao and Xu, Runxin and Song, Junxiao and Bi, Xiao and Zhang, Haowei and Zhang, Mingchuan and Li, YK and Wu, Yang and others},
  journal={arXiv preprint arXiv:2402.03300},
  year={2024}
}

@article{rafailov2023direct,
  title={Direct preference optimization: Your language model is secretly a reward model},
  author={Rafailov, Rafael and Sharma, Archit and Mitchell, Eric and Manning, Christopher D and Ermon, Stefano and Finn, Chelsea},
  journal={Advances in neural information processing systems},
  volume={36},
  pages={53728--53741},
  year={2023}
}

@article{luo2024robustft,
  title={Robustft: Robust supervised fine-tuning for large language models under noisy response},
  author={Luo, Junyu and Luo, Xiao and Ding, Kaize and Yuan, Jingyang and Xiao, Zhiping and Zhang, Ming},
  journal={arXiv preprint arXiv:2412.14922},
  year={2024}
}

@article{schmirler2024fine,
  title={Fine-tuning protein language models boosts predictions across diverse tasks},
  author={Schmirler, Robert and Heinzinger, Michael and Rost, Burkhard},
  journal={Nature Communications},
  volume={15},
  number={1},
  pages={7407},
  year={2024},
  publisher={Nature Publishing Group UK London}
}

@article{sledzieski2024democratizing,
  title={Democratizing protein language models with parameter-efficient fine-tuning},
  author={Sledzieski, Samuel and Kshirsagar, Meghana and Baek, Minkyung and Dodhia, Rahul and Lavista Ferres, Juan and Berger, Bonnie},
  journal={Proceedings of the National Academy of Sciences},
  volume={121},
  number={26},
  pages={e2405840121},
  year={2024},
  publisher={National Academy of Sciences}
}

@article{Arnold2017DirectedEB,
  title={Directed Evolution: Bringing New Chemistry to Life},
  author={Frances H. Arnold},
  journal={Angewandte Chemie (International Ed. in English)},
  year={2017},
  volume={57},
  pages={4143 - 4148},
  url={https://api.semanticscholar.org/CorpusID:4727910}
}

@article{wang2025supervised,
  title={Supervised fine-tuning of pre-trained antibody language models improves antigen specificity prediction},
  author={Wang, Meng and Patsenker, Jonathan and Li, Henry and Kluger, Yuval and Kleinstein, Steven H},
  journal={PLOS Computational Biology},
  volume={21},
  number={3},
  pages={e1012153},
  year={2025},
  publisher={Public Library of Science San Francisco, CA USA}
}

@article{bhethanabotlacapturing,
  title={Capturing Protein Dynamics: Encoding Temporal and Spatial Dynamics from Molecular Dynamics Simulations},
  author={Bhethanabotla, Vignesh C and Tavakoli, Mohammadamin and Annandkumar, Anima and Goddard III, William A}
}

@article{lambert2025sequence,
  title={Sequence-Based Generative AI-Guided Design of Versatile Tryptophan Synthases},
  author={Lambert, Th{\'e}ophile and Tavakoli, Amin and Dharuman, Gautham and Yang, Jason and Bhethanabotla, Vignesh and Kaur, Sukhvinder and Hill, Matthew and Ramanathan, Arvind and Anandkumar, Anima and Arnold, Frances H},
  journal={bioRxiv},
  pages={2025--08},
  year={2025},
  publisher={Cold Spring Harbor Laboratory}
}



\section{Appendix}

\subsection{Enzyme Family}

We study the enzyme tryptophan synthase, a multi-subunit complex that catalyzes the final steps of tryptophan biosynthesis. The enzyme functions as a heterotetramer composed of two $\alpha$ and two $\beta$ subunits. In this work, we focus on the $\beta$ subunit (TrpB), which catalyzes the condensation of indole and L-serine to form L-tryptophan.

Catalytic activity of TrpB is typically assayed by quantifying L-tryptophan production from indole and L-serine, using methods such as spectrophotometry \cite{gore2000spectrophotometry} or high-performance liquid chromatography (HPLC) \cite{tsao2003polyphenolic}. These assays are labor-intensive and time-consuming, requiring days to validate only a small fraction of candidates. This underscores the challenge of obtaining high-quality data for SFT in PLMs compared to the abundant, quickly human-annotated data available for LLMs. It also highlights the critical role of computational annotations and PLM sampling quality.

From a mechanistic perspective, functional TrpBs must satisfy several biochemical constraints: (1) proper docking of the pyridoxal 5'-phosphate (PLP) cofactor within the active site; (2) sufficient structural stability to support catalysis; (3) sequence length consistent with natural variants; and (4) conservation and accessibility of the catalytic lysine residue required for PLP binding and enzymatic turnover. A generative model for TrpB sequences must therefore produce candidates that meet these structural and functional requirements.  These biochemical criteria inform the design of our filtering pipeline and supervised fine-tuning strategy, ensuring that the model is guided toward generating functionally plausible enzymes.

\subsubsection{Reference Set}

To assess the novelty and plausibility of the generated protein sequences, it is essential to compare them against a comprehensive set of active and functional variants (e.g., natural variants). The reference set provides a baseline for measuring sequence identity, structural similarity, and conservation of activity features, enabling systematic filtering and validation of model outputs.

We assembled 57,250 natural TrpB sequences as the reference set. 30,000 were retrieved from the Bacterial and Viral Bioinformatics Resource Center (BV-BRC) database using EC number 4.2.1.20 \cite{pickett2012vipr}, with low-quality entries excluded by requiring complete genomes (<5\% contamination, <50 contigs). Since this EC class includes both subunits, only features annotated as the beta subunit were retained. The remaining sequences were collected from \cite{uniprot2023} using the gene name trpB and constrained to lengths between 200 and 600 amino acids.

\end{document}